\documentclass{article}
\usepackage{nips00e}
\usepackage{array,amsgen,amssymb,amsopn,amsmath}
\usepackage{amsthm}
\usepackage[pdftex]{graphicx}
\usepackage[square,sort,comma,numbers]{natbib}
\usepackage[latin1]{inputenc}
\usepackage{verbatim}
\usepackage{bbm}

\newcommand\given{\medspace|\medspace}
\newcommand\argmax{\operatornamewithlimits{argmax}}

 \title{Probabilistic Transformers}
\author{ Javier R. Movellan \&  Prasad Gabbur \\Apple}

\empty

\begin{document}
\maketitle

\begin{abstract}
We 
show that Transformers are Maximum Posterior Probability estimators for  Mixtures of Gaussian Models. This brings a probabilistic  point of view  to Transformers and suggests extensions  to inference-time model adaptation and to other probabilistic cases. 
\end{abstract}

\newpage
We adopt the  interpretation of Transformers \cite{cc,bb} as knowledge systems  consisting of $n$ memory units that can be independently queried to get answers. The queries are $d$-dimensional vectors,  $q \in R^d$ and the answers are vectors $v\in R^m$ called values. The queries and the corresponding values may depend on an input vector $x$. For example, each unit may represent  a pixel in an image, $x$. A query, $q$, could be a vector that  requests the semantic category of that pixel, and the answer, $v$,  will depend on the image $x$. Probabilistic Transformers provide a probabilistic interpretation to Standard Transformers \cite{cc,bb}  and suggest new algorithms and extensions. A Probabilistic Transformer is defined  by the  joint distribution of queries $q$  and values $v$ given inputs $x$. This distributions factorizes as follows: 
\begin{align}
p(q_{1:n}, v_{1:n}, y_{1:n} | x) = \Pi_{i=1}^n p_i(q_i,v_i \given x )
\end{align}
where $x$ is the conditioning input vector, $q_{1:n} = \{ q_1, \cdots, q_n\}$, $v_{1:n} = \{v_1,\cdots, v_n\}$, and   $q_i,v_i \in R^d, x_i \in R^m$ are the query, value vectors for unit $i$.  Here  $p_i(q,v \given x )$ is the joint distribution of queries received and values produced by unit $i$ given an input vector $x$. This joint distribution is a probabilistic mixture model:
\begin{align}
p_i(q,v \given x ) = \sum_{j=1}^n p_i(q,v,u_j\given x ) = \sum_i \pi_{i,j} (x) \:  p_i(q,v\given u_j,x)  
\end{align}
where $\pi_{i,j}(x)$ is the probability of activating unit $j$ when unit $i$ is queried, $p_i(q,v \given u_j,x )$ is the probability that unit $j$  generates the pair  $q,v$  given the input $x$.

\paragraph{Optimal Value Inference:}
Gven the input $x$ and a query $q$ to unit $i$ we want to infer the most probable value  $\hat v$
\begin{align}
\hat v = \argmax_v p_i(v| q,x) 
\end{align}
To this end  we use an Expectation  Maximization (EM) approach: We start with an initial estimate $v^0$ of the most probable value and iterate to optimize the standard EM auxiliary function $Q_i$. Given  the latest  known estimate $v^t$ we look for  a new estimate $v^{t+1}$ that increases $Q_i$. Maximizing $Q_i$ with respect to $v^{t+1}$ guarantees local maximization of $p_i(v|q,x)$. Let 
\begin{align}
Q_i(v^t,v^{t+1}\given x) &= \sum_j w_{i,j}^t \log p_i(u_j,q,v^{t+1}\given x) \\
w_{i,j}^t &= p_i(u_j\given q,v^t, x) =  \frac{ \pi_{i,u}(x)  \:p_i(q,v^t \given u_j, x) }{\sum_j  \pi_{i,j}(x)  \:p_i(q, v^t\given u_j, x) }
\end{align}
The $n\times n$ matrix $w$ corresponds to the {\em Attention } matrix in Standard Transformers. Here $w_{i,j}$ controls the influence of unit $j$ has on the optimal value estimate for unit $i$.  Taking the gradient with respect to $v^{t+1}$ and setting it to zero we get the EM maximization equation
\begin{align}\label{eqn:general}
	\nabla_{v^{t+1}} Q_i(v^t,v^{t+1} \given x ) = \sum_j w_{i,j} \nabla_{v^{t+1}} \log p_i(q, v^{t+1}, u_j \given x) =0 
\end{align}
where  $\nabla_{v^{t+1}}  \log p_u(q, v_{t+1}, u \given x)$ is the Fisher Score for unit $j$ with respect to $v^{t+1}$.

\paragraph{Relation to Standard Transformers:}
Here we show that Standard Transformers \cite{bb,cc} solve  \eqref{eqn:general}  when the observation model is a mixture of Gaussians with constraints  on the prior probabilities and covariance matrices of the mixtures. Let

\begin{align}
 &p_i(q,v \given u_j,x)  = p_i(q\given u_j,x) \: p_i(v \given u_j,x) \\
&p_i(q \given u_j,x) = \Big( \frac{\alpha_i(x)}{2 \pi}\Big)^{d/2}\; e^{-\frac{\alpha_i(x)}{2}  \| q - \xi_j(x)\|^2 }\\
&p_i(v \given j,x) = \Big( \frac{\beta_i(x)}{2 \pi}\Big)^{m/2}\; e^{-\frac{\beta_i(x)}{2}  \| v - \mu_j(x)\|^2 }
\end{align}
 where $\alpha_i(x), \beta_i(x) > 0$ are precision parameters,  $\xi_j(x) \in R^d$, $\mu_j(x) \in R^m$ are the key and expected value parameters for unit $j$ given the input vector $x$.  Note the dependency of $p_i(q,v \given u,x)$ on $x$ is through the fact that the parameters $\alpha_j, \beta_j, \pi_{i,j}$, $\xi_j, \mu_j$ are a function of $x$.   To simplify the presentation hereafter we treat  $x$  as a fixed input vector and leave the dependency on $x$  implicit in our notation.  Note
 \begin{align}
&p_i(q,v) =  \sum_j \frac{1}{z_j} \:\pi_{i,j} e^{-\frac{\alpha_j}{2}  \| q - \xi_j\|^2 } e^{-\frac{\beta_j}{2}  \| v - \mu_j\|^2 }\\
&z_j =\Big( \frac{2 \pi}{\alpha_j}\Big)^{d/2} \Big( \frac{2 \pi}{\beta_j}\Big)^{m/2}
\end{align}
and the Fisher score takes the following form
\begin{align}
	\nabla_{v^{t+1}} \log p_i(q, v^{t+1}, u_j)  =\beta_j   (\mu_j-v^{t+1})
\end{align}
Thus
\begin{align}
	\nabla_{v^{t+1}} Q_i(v^t, v^{t+1}) = \sum_j w_{i,j}^t \beta_j  (\mu_j -v^{t+1}) =0
\end{align}
and the EM maximization equation becomes as follows:
\begin{align}\label{eqn:gauss}
& v^{t+1}  =  \sum_j  w_{i,j}^t\: \mu_j \\
& w_{i,j}^t  = \frac{ 
\pi_{i,j} \: \beta_j\:e^{-\frac{\alpha_j}{2}  \| q - \xi_j\|^2 } \:e^{-\frac{\beta_j}{2}  \| v^t - \mu_j\|^2 }}
{ 
\sum_j \pi_{i,j} \beta_j  \: e^{-\frac{\alpha_j}{2}  \| q - \xi_j\|^2 } \:e^{-\frac{\beta_j}{2}  \|  v^t  - \mu_j\|^2 }}
\end{align}

To get the Standard Transformer equation we constrain the precision parameters to be equal across units: $\alpha_1 = \cdots = \alpha_n = \alpha$,  $\beta_1 = \cdots = \beta_n = \beta$, and  we  link the  priors of each unit to the length of the key and expected value vectors 
\begin{align}
\pi_{i,j}  &=\frac{1}{k} e^{\frac{\alpha}{2} \| \xi_j\|^2 }  e^{\frac{\beta}{2} \| \mu_j\|^2 } \\
k&= \sum_j e^{\frac{\alpha}{2}  \| \xi_j\|^2 }  e^{\frac{\beta}{2} \| \mu_j\|^2 }  
\end{align}
Note that this makes $w_{i,j}$  independent of $i$ (permutation equivariant)   and simplifies the optimal inference equation  as follows:
\begin{align}\label{eqn:optimalv}
&v^{t+1}  = \sum_j w_{i,j} \mu_j\\
&w_{i,j}^t = \frac{ e^{\alpha \xi_j' q} \;e^{\beta   \mu_j'v_t }  }
{\sum_j e^{\alpha\xi_j' q}\;  \;e^{\beta  \mu_j' v_t }} 
\end{align}
As $\beta \to 0$ we obtain the Standard Transformer equation:
\begin{align}\label{eqn:optimalv}
&v^{t+1} = \sum_j w_{i,j} \mu_j\\
&w_{i,j}^t = \frac{ e^{\alpha \xi_j' q}   }
{\sum_j e^{\alpha\xi_j' q}}
\end{align}
Note in this case $w_{i,j}^t$ is no longer a function of $t$ and thus only one EM iteration is needed.  In conclusion, we have shown that Standard Transformers  \cite{cc,bb} can be seen as performing MAP inference under a special case of the Probabilistic Transformer model. 
\paragraph{Off-line  Supervised Learning:}
As is commonly done in standard Transformers, the relationship between the input $x$ and the  parameters: $\pi(x), \xi(x), \mu(x)$ can be parameterized, embedded in a deep network and trained off-line using Stochastic Gradient Descent. 

\paragraph{On line Unsupervised Key Learning:}
At inference time we typically receive an input vector $x$ and a collection of queries, one per unit: $q_{1:n} = \{ q_1, \cdots, q_n\}$. Prior to inferring the most probable values for each unit it is possible to adapt the key vectors, $\xi_{1:n} = \{ \xi_1,\cdots, \xi_n\}$ in an unsupervised manner. For each unit $i$ we want
\begin{align}
&\hat v_i = \argmax_v p_i (v \given q_{1:n})  
\end{align}
To this end we adopt a MAP approach
\begin{align}
&p_i( v \given  q_{1:n} ) = \int p(\xi_{1:n}  \given q_{1:n} ) p_i (v \given q_i, \xi_{1:n}) d \xi_{1:n}  \approx  p_i( v \given q_i,\hat \xi_{1:n}) \\ 
&\hat \xi_{1:n} = \argmax_{ \xi_{1:n}}   p (\xi_{1:n}  \given q_{1:n})\label{eqn:maxxi}\\
&\hat v_i = \argmax_v p_i (v \given q_i, \hat  \xi_{1:n}  ) \label{eqn:maxv1}
\end{align}
To solve \eqref{eqn:maxxi} we use an iterative EM approach. The initial key parameters $\xi^0_{1:n}$ are provided by the pre-trained model. To avoid overfitting to the current query vectors we use a Gaussian prior  centered on the key parameters provided by the pre-trained network, i.e., $\xi^0_{1:n}$. 
\begin{align}
&Q(\xi_{1:n}^t , \xi_{1:n}^{t+1} ) = \log p(\xi^{t+1}_{1:n}) + \sum_{i=1}^n \sum_{j=1}^n  w_{i,j}^t \log p_j( q_i, u_j \given \xi_j^{t+1})  \\
& w_{i,j}^t = p_i(u_j \given q_i, \xi^t_{1:n} )  = \frac{\pi_{i,j} \: p(q_i \given u_j, \xi^t_j )}{\sum_{k=1}^n \pi_{i,k} \:  p(q_i\given u_k, \xi_k^t)}
\end{align}
\begin{align}
\nabla_{\xi_k^{t+1} } Q(\xi_{1:n}^t , \xi_{1:n}^{t+1} ) = \theta_\xi ( \xi_k^0- \xi^{t+1}_k) + \sum_{i=1}^s w_{i,k}^t  \alpha_k (q_i - \xi^{t+1}_k )
\end{align}
Where $\theta_{\xi} >0 $ is the precision of the Gaussian prior over keys.  Setting the gradients to zero and solving for $\xi_k^{t+1}$ we get the EM update equation
\begin{align}
\xi_k^{t+1} = \frac{\theta_{\xi} \xi^0_k  + \alpha_k \sum_{i=1}^n w_{i,k}^t q_i }{ \theta_\xi + \alpha_k \sum_{i=1}^n w_{i,k}^t}
\end{align}
We can also adapt the $\alpha$ precision parameters. To avoid overfit we use a Gamma prior with parameters $\theta_{\alpha,1}, \theta_{\alpha,2}$. In this case the EM update equations look as follow

\begin{align}
\alpha_k^{t+1} = \frac{\theta_{\alpha,1}  + d/2 \sum_{i=1}^n w_{i,k}^t -1}{ \theta_{\alpha,2} +\sum_{i=1}^n w_{i,k}^t\: \frac{1}{2} \| q_i - \xi_k\|^2 }
\end{align}

\paragraph{On Line Belief Propagation:}
In some applications additional   information is obtained at inference time that could be used to improve the accuracy of the inference.  For example in Interactive Semantic Segmentation  the memory units correspond to pixels, and the values  correspond to the semantic category of that pixel.  A deep network may produce queries $q_{1:n}$ for all the units and the transformer gets the most probable values for each pixel. The transformers may make mistakes in some pixels and the human annotator may then provide the correct values for a subset of those pixels. We want the new information about the correct values for some pixels to propagate to all the other pixels. Here we propose an approach for this type of belief propagation within the framework of Probabilistic Transformers. Suppose  the annotator has provided the correct values  for the first $s<n$ units.  We want for this information to improve the inference about the value for all the other units $i > s$.  Within our framework we want 
\begin{align}
\hat v_i = \argmax_v p_i (v \given q_i, q_{1:n}, v_{1:s} ), \: \text{for $s<i<=n$}
\end{align}
To this end we adopt a MAP approach. Let $\lambda$ represent network parameters, e.g., $\pi,\xi,\mu, \alpha, \beta$. We note that
\begin{align}
&p_i( v \given  q_{1:n}, v_{1:s} ) = \int p(\lambda\given q_{1:n}, v_{1:s}  ) p_i(v \given q_i, \lambda) d \lambda   \approx  p_i( v \given q_i,\hat \lambda)\\
&\hat \lambda = \argmax_\lambda p (\lambda \given q_{1:n}, v_{1:s})\label{eqn:lambda}\\
&\hat v_i = \argmax_v p_i(v \given q_i, \hat \lambda)
\end{align}
As in the previous Section we solve \eqref{eqn:lambda} using an EM approach. For example for optimizing the expected value for unit $k$, we start with the initial vector $\mu^0_k$ provided by the pre-trained model. At each EM iteration we have an estimate $\mu_k^t$ and update to a better estimate $\mu_k^{t+1}$. Following a similar derivation as in the previous section, we get the following update equation
\begin{align}
&\mu_k^{t+1} = \frac{\theta_{\mu} \mu^0_k  + \beta_k \sum_{i=1}^s w_{i,k}^t v_i }{ \theta_\mu+ \beta_k \sum_{i=1}^s w_{i,k}^t}\\
& w_{i,k}^t = p_i(u_k \given q_i,v_i, \mu^t_{1:n} )  = \frac{\pi_{i,k} \: p(q_i \given u_k, \xi_k )\:p(v_i \given u_k, \mu^t_k )}{\sum_{j=1}^n \pi_{i,k} \: p(q_i \given u_j, \xi_j )\:p(v_i \given u_j, \mu_j^t )}\end{align}
where $\theta_\mu$ is the precision for the Gaussian prior over values. In a similar fashion we can derive update equations for the $\beta$ and $\pi$ parameters.

\begin{align}
&\beta_k^{t+1} = \frac{\theta_{\beta,1}  + d/2 \sum_{i=1}^s w_{i,k}^t -1}{ \theta_{\beta,2} +\frac{1}{2} \sum_{i=1}^s w_{i,k}^t \| v_i - \mu_k\|^2 }\\
&\pi_{i,k}^{t+1} = \frac{ w_{i,k}^t + \theta_{\pi,i,k} -1}{\sum_k w_{i,k}^t + \theta_{\pi,i,k} -1}
\end{align}
where $\theta_{\beta,1}, \theta_{\beta,2}$ are the parameters for a Gamma prior distribution over $\beta_k$, and $\theta_{\pi,i,k}$ are Dirichlet prior parameters over $\pi_{i,k}$. 
\paragraph{Combining Off-line learning and On-line Adaptation} 
The Inference-Time  adaptation of parameters is differentiable, so it can be included as part of the overall algorithm trained via Gradient descent and used to learn the parameters of the prior distributions over $\xi,\mu, \alpha, \beta,\pi$.

\end{document}